\documentstyle[aclap]{article}

\newcounter{example}

\author{
Diane J. Litman	\\
AT\&T Bell Laboratories\\
600 Mountain Avenue\\
Murray Hill, NJ  07974 \\
diane@research.att.com \\
\And
Rebecca J. Passonneau\thanks{Bellcore did not support the second author's
work.}\\
Bellcore \\
445 South Street \\
Morristown, NJ 07960  \\
beck@bellcore.com \\
}

\title{\vspace{-0.5in}Combining Multiple Knowledge Sources for
Discourse Segmentation}

\begin{document}

\maketitle

\begin{abstract}
We predict discourse segment boundaries from linguistic
features of utterances, using a corpus of spoken narratives as data.
We present two methods for developing segmentation algorithms from
training data: hand tuning and machine learning.  When multiple types
of features are used, results approach human
performance on an independent test set (both
methods), and using cross-validation (machine learning).
\end{abstract}

\section{Introduction}

Many have argued that discourse has a global structure above
the level of individual utterances,
and that linguistic phenomena
like prosody,
cue phrases, and
nominal reference
are partly conditioned by and reflect this structure
(cf.~\cite{grosz&hirschberg92,gs86,hirschberg&grosz92,hl93}
{}~\cite{hirschberg&pierrehumbert86,hobbs79,lascarides&oberlander92,linde79}
{}~\cite{mann&thompson88,polanyi88,Reichman85,webber91}).
However, an obstacle
to exploiting the relation between global structure and
linguistic devices in natural language systems
is that there is too little data
about how they constrain one another.
We have been engaged in a study addressing this gap.
In previous work~\cite{passonneau&litman93}, we reported on a method for
empirically
validating global discourse units, and on our evaluation of
algorithms to identify these units.  We found significant agreement
among naive subjects on a discourse segmentation task, which
suggests that global discourse units have some objective reality.
However, we also found poor correlation of
three untuned algorithms (based on features of
referential noun phrases, cue words, and pauses, respectively)
with the subjects' segmentations.

In this paper, we discuss two methods for developing segmentation
algorithms using multiple knowledge sources.
In section~\ref{background}, we give
a brief overview of related work and summarize our previous results.
In section~\ref{method}, we discuss how linguistic features are
coded and describe our evaluation.
In section~\ref{handTun}, we present
our analysis of the errors
made by the best performing untuned algorithm, and a new algorithm that
relies on enriched input features and multiple
knowledge sources.
In section~\ref{mlearn}, we discuss our use of machine learning tools to
automatically construct decision trees for segmentation
from a large set of input features.
Both the hand tuned and automatically derived algorithms improve
over our previous algorithms.
The primary benefit of the hand tuning is to identify new input features
for improving performance.  Machine learning tools make it
convenient to perform numerous experiments, to use large feature
sets, and to evaluate results using
cross-validation.
We discuss the significance of our results and briefly compare the two
methods in section~\ref{conclusion}.

\section{Discourse Segmentation}
\label{background}

\subsection{Related Work}
\label{relWk}

Segmentation has played a significant role in much work on discourse.
The linguistic structure of Grosz and Sidner's~\shortcite{gs86}
tri-partite discourse model
consists of multi-utterance segments
whose hierarchical relations are isomorphic with intentional structure.
In other work (e.g.,~\cite{hobbs79,polanyi88}),
segmental structure is an artifact of coherence relations among
utterances, and few if any specific claims are made regarding
segmental structure per se.
Rhetorical Structure
Theory (RST)~\cite{mann&thompson88} is another tradition of defining
relations among utterances, and informs much work in generation.
In addition, recent
work~\cite{moore&paris93,moore&pollack92} has addressed the
integration of intentions and rhetorical relations.  Although all of
these approaches have involved detailed analyses of individual
discourses or representative corpora, we believe there is a need
for more rigorous empirical studies.

Researchers
have begun to investigate the ability of humans to agree with one
another on segmentation, and to propose methodologies for quantifying
their findings.  Several studies have used expert
coders to locally and globally
structure spoken discourse according to the model of Grosz and
Sidner~\shortcite{gs86}, including~\cite{grosz&hirschberg92}
{}~\cite{hirschberg&grosz92,Nakatani95,stifleman95}.
Hearst~\shortcite{hearst94} asked
subjects to place boundaries between paragraphs of expository texts,
to indicate topic changes.  Moser and Moore~\shortcite{Moser&Moore95} had
an expert coder assign segments and various segment features and
relations based on RST.
To quantify their findings,
these studies use notions of agreement~\cite{GCY,Moser&Moore95} and/or
reliability~\cite{passonneau&litman93,passonneau&litman95,Isard&Carletta95}.

By asking subjects to segment discourse using a non-linguistic
criterion, the correlation of linguistic devices with independently
derived segments can then be investigated in a way that avoids
circularity.   Together,~\cite{grosz&hirschberg92,hirschberg&grosz92}
{}~\cite{Nakatani95} comprise an ongoing study using
three corpora: professionally read AP news stories, spontaneous
narrative, and read and spontaneous versions of task-oriented monologues.
Discourse structures are derived from subjects' segmentations,
then statistical measures are used to characterize these
structures in terms of acoustic-prosodic features.
Grosz and Hirschberg's work
also used the classification and regression tree system
CART~\cite{brieman84}
to automatically construct and evaluate decision
trees for classifying
aspects of discourse structure from intonational feature values.
Morris and
Hirst~\shortcite{morris&hirst91} structured a set of magazine texts using the
theory of~\cite{gs86}, developed a thesaurus-based lexical cohesion
algorithm to segment text, then qualitatively compared their
segmentations with the results.
Hearst~\shortcite{hearst94}
presented two implemented segmentation algorithms based on term
repetition, and compared the boundaries produced to the boundaries
marked by at least 3 of 7 subjects, using information retrieval metrics.
Kozima~\shortcite{kozima93} had 16 subjects segment a simplified short story,
developed an algorithm based on lexical cohesion, and qualitatively
compared the results.  Reynar~\shortcite{Reynar94} proposed an algorithm
based on lexical cohesion in conjunction with a graphical technique, and
used information retrieval metrics to evaluate the algorithm's
performance in locating boundaries between concatenated news
articles.

\subsection{Our Previous Results}
\label{prev}



We have been investigating a corpus of monologues collected
and transcribed by Chafe~\shortcite{chafe80}, known as the Pear stories.
As reported in~\cite{passonneau&litman93},
we first investigated whether units
of global structure consisting of sequences of utterances could be
reliably identified by naive subjects.
We analyzed linear segmentations of 20 narratives
performed by naive subjects (7 new subjects per narrative), where
speaker intention was the segment criterion.
Subjects
were given transcripts, asked to place a new segment boundary between
lines (prosodic phrases)\footnote{We used Chafe's~\shortcite{chafe80}
prosodic analysis.}
wherever the speaker had a new communicative goal, and to briefly
describe the completed segment.  Subjects were free to assign
any number of boundaries.  The qualitative results were that
segments varied in size from 1 to 49 phrases in length (Avg.=5.9),
and the rate at which subjects assigned boundaries ranged from
5.5\% to 41.3\%.
Despite this variation, we found statistically significant
agreement among subjects across all narratives
on location of segment boundaries ($.114 \;
x \; 10^{-6} \; < \; p \; < \; .6 \; x \; 10^{-9}$).

We then looked at the predictive power of
linguistic cues for identifying the segment boundaries agreed upon by a
significant number of subjects.  We used three distinct algorithms
based on the distribution of referential noun phrases, cue
words, and pauses, respectively.  Each algorithm
(NP-A, CUE-A, PAUSE-A) was designed to
replicate the subjects' segmentation task
(break up a narrative into contiguous segments, with segment
breaks falling between prosodic phrases).
NP-A used three
features,
while CUE-A and PAUSE-A
each made use of a single feature.
The features are a subset of those described in section~\ref{method}.

To evaluate how well an algorithm predicted segmental structure, we
used the information retrieval (IR) metrics described in
section~\ref{method}.  As reported in~\cite{passonneau&litman95}, we
also evaluated a simple additive method for combining algorithms in
which a boundary is proposed if each separate algorithm proposes a
boundary.
We tested all pairwise combinations,
and the combination of all three
algorithms.
No algorithm or
combination of algorithms performed as well as humans.
NP-A performed better than the other unimodal algorithms, and a
combination of NP-A and PAUSE-A performed best.  We felt
that significant improvements could be gained by
combining the input features in more complex ways
rather than by simply combining the outputs of independent algorithms.

\section{Methodology}
\label{method}

\subsection{Boundary Classification}

We represent each narrative in our corpus as a sequence of potential
boundary sites, which occur between prosodic phrases.
We classify a potential boundary site as {\it boundary} if it was
identified as such by at least 3 of the 7 subjects in our earlier
study.  Otherwise it is classified as {\it non-boundary}.
Agreement
among subjects on boundaries was significant at below the .02\%
level for values of j $\geq$ 3, where j is the number of subjects (1
to 7), on all 20 narratives.\footnote{We previously used
agreement by 4 subjects as the threshold for boundaries; for j $\geq$
4, agreement was significant at the .01\%
level.~\cite{passonneau&litman93}}

Fig.~\ref{excerpt6} shows a typical segmentation of one of the narratives in
our corpus.  Each line corresponds to a prosodic phrase, and each
space between the lines corresponds to a potential boundary
site.  The bracketed numbers will be explained below.  The boxes in
the figure show the subjects' responses at each
potential boundary site, and the resulting boundary classification.
Only 2 of the 7 possible boundary
sites are classified as {\it boundary}.
\begin{figure}
{\scriptsize
\begin{tabbing}
ssss \= sssssssssssssssassssssssssssssssssssssssssssssssss \kill
..Because he's looking at the girl.   \\
\hspace{.25in} \fbox{1 SUBJECT \hspace{.03in}  {\it (non-boundary})} \\
{[}.75] Falls over, \\
\hspace{.25in} \fbox{5 SUBJECTS \hspace{.2in} ({\it boundary})} \\
{[}1.35] uh there's no conversation in this movie.  \\
\hspace{.25in} \fbox{0 SUBJECTS {\it (non-boundary})} \\
{[}.6] There's sounds,  \\
\hspace{.25in} \fbox{0 SUBJECTS {\it (non-boundary})} \\
you know,  \\
\hspace{.25in} \fbox{0 SUBJECTS {\it (non-boundary})} \\
like the birds and stuff,  \\
\hspace{.25in} \fbox{0 SUBJECTS {\it (non-boundary})} \\
but there.. the humans beings in it don't say anything.  \\
\hspace{.25in} \fbox{7 SUBJECTS \hspace{.2in}  ({\it boundary})} \\
{[}1.0] He falls over,
\end{tabbing}
}
\caption{\label{excerpt6} Excerpt from narr. 6, with boundaries.}
\end{figure}

\subsection{Coding of Linguistic Features}

Given a narrative of n prosodic phrases, the
n-1 potential boundary sites are between each pair of
prosodic phrases P$_{i}$
and P$_{i+1}$, i from 1 to n-1.
Each potential boundary site
in our corpus is coded using the set of linguistic features
shown in Fig.~\ref{features}.

\begin{figure}
{\scriptsize
\begin{center}
\begin{itemize}
\item
{\bf Prosodic Features}
\begin{itemize}
\item
before:+sentence.final.contour,-sentence.final.contour
\item
after: +sentence.final.contour,-sentence.final.contour.
\item
pause: true, false.
\item
duration: continuous.
\end{itemize}
\item
{\bf Cue Phrase Features}
\begin{itemize}
\item
cue$_{1}$: true, false.
\item
word$_{1}$: also, and, anyway, basically, because, but, finally, first, like,
meanwhile, no, now, oh, okay, only, or, see, so, then, well, where, NA.
\item
cue$_{2}$: true, false.
\item
word$_{2}$: and, anyway, because, boy, but, now, okay, or, right, so, still,
then, NA.
\end{itemize}
\item
{\bf Noun Phrase Features}
\begin{itemize}
\item
coref: +coref, -coref, NA.
\item
infer: +infer, -infer, NA.
\item
global.pro: +global.pro, -global.pro, NA.
\end{itemize}
\item
{\bf Combined Feature}
\begin{itemize}
\item
cue-prosody: complex, true, false.
\end{itemize}
\end{itemize}
\caption{\label{features} Features and their potential values.}
\end{center}
}
\end{figure}

Values for the prosodic features are obtained by automatic
analysis of the transcripts, whose conventions
are defined in~\cite{chafe80} and illustrated in Fig.~\ref{excerpt6}:
``.'' and ``?''  indicate sentence-final
intonational contours; ``,'' indicates phrase-final but not sentence final
intonation;
``[X]'' indicates a pause lasting X
seconds;  ``..'' indicates a break in timing too short to
be measured.
The features {\it before} and {\it after}
depend on the final punctuation of the phrases P$_{i}$ and
P$_{i+1}$, respectively.
The value is `+sentence.final.contour' if ``.''
or ``?'', `-sentence.final.contour' if ``,''.  {\it Pause}
is assigned `true' if P$_{i+1}$ begins with [X], `false' otherwise.
{\it Duration} is assigned X if {\it pause} is `true', 0
otherwise.

The cue phrase features are also obtained by automatic analysis of the
transcripts.  {\it Cue$_{1}$} is assigned `true' if the first lexical
item in P$_{i+1}$ is a member of the set of cue words summarized
in~\cite{hl93}.  {\it Word$_{1}$} is assigned this lexical item if
{\it cue$_{1}$} is true, `NA' (not applicable)
otherwise.\footnote{The cue phrases that occur in the corpus are
shown as potential values in Fig.~\ref{features}.}
{\it Cue$_{2}$} is assigned `true' if {\it
cue$_{1}$} is true and the second lexical item is also a cue word.
{\it Word$_{2}$} is assigned the second lexical item if {\it
cue$_{2}$} is true, `NA' otherwise.

Two of the noun phrase (NP) features are
hand-coded, along with functionally independent clauses (FICs),
following~\cite{passonneau94cod}.
The two authors coded independently and merged
their results.
The third feature, {\it global.pro}, is computed from the hand coding.
FICs are tensed clauses that are neither
verb arguments nor restrictive relatives.
If a new FIC
(C$_{j}$) begins in prosodic phrase P$_{i+1}$, then NPs in C$_{j}$ are
compared with NPs in previous clauses and the feature values assigned as
follows\footnote{The NP algorithm can assign multiple boundaries
within one prosodic phrase if the phrase contains multiple clauses;
these very rare cases are
normalized~\cite{passonneau&litman93}.}:
\begin{enumerate}
\item
{\it coref} = `+coref' if C$_{j}$ contains an NP that corefers with an NP in
C$_{j-1}$; else {\it coref} = `-coref'
\item
{\it infer} = `+infer' if C$_{j}$ contains an NP whose referent can be
inferred from an NP in C$_{j-1}$ on the basis of a pre-defined set of inference
relations; else {\it infer} = `-infer'
\item
{\it global.pro} = `+global.pro' if C$_{j}$ contains a definite pronoun whose
referent is mentioned in a previous clause up to the last boundary
assigned by the algorithm; else {\it global.pro} = `-global.pro'
\end{enumerate}
If a new FIC is not initiated in P$_{i+1}$,
values for all three features are `NA'.

\begin{figure*}[t]
{\scriptsize
..Because he$_{i}$'s looking at the girl. \\
{[}.75] (ZERO-PRONOUN$_{i}$) Falls over, \\
\begin{center}
\begin{tabular}{r r r r r r r r r r r r}
before &after &pause &duration &cue$_{1}$ &word$_{1}$ &cue$_{2}$ &word$_{2}$
&coref &infer &global.pro &cue-prosody \\ \hline
+s.f.c &-s.f.c &true &.75 &false &NA &false &NA & + & - & + & true\\
\end{tabular}
\caption{\label{coding} Example feature coding of a potential boundary site.}
\end{center}}
\end{figure*}

{\it Cue-prosody}, which encodes a combination of
prosodic and cue word features, was motivated by
an analysis of IR errors on our training data, as
described in section~\ref{handTun}.
{\it Cue-prosody} is
`complex' if:
\begin{enumerate}
\item {\it before} = `+sentence.final.contour'
\item {\it pause} = `true'
\item And either:
\begin{enumerate}
\item {\it cue$_{1}$} = `true', {\it word$_{1}$} $\neq$ `and'
\item {\it cue$_{1}$} = `true', {\it word$_{1}$} = `and',
{\it cue$_{2}$} = `true', {\it word$_{2}$} $\neq$ `and'
\end{enumerate}
\end{enumerate}
Else, {\it cue-prosody} has the same values
as {\it pause}.

Fig.~\ref{coding} illustrates how the first boundary site in
Fig.~\ref{excerpt6} would be coded using the features in
Fig.~\ref{features}.

The prosodic and cue phrase features were motivated by previous
results in the literature.  For example, phrases beginning discourse
segments were correlated with preceding pause duration
in~\cite{grosz&hirschberg92,hirschberg&grosz92}.  These and other
studies (e.g.,~\cite{hl93}) also found it useful to distinguish
between sentence and non-sentence final intonational contours.
Initial phrase position was correlated with discourse signaling uses
of cue words in~\cite{hl93}; a potential correlation between discourse
signaling uses of cue words and adjacency patterns between cue words
was also suggested.  Finally,~\cite{litman94} found that treating cue
phrases individually rather than as a class enhanced the results
of~\cite{hl93}.

Passonneau~\shortcite{passonneau95} examined some of the
few claims relating discourse anaphoric noun phrases to global
discourse structure in the Pear corpus.
Results included an absence of correlation of
segmental structure with centering~\cite{gjw83,kameyama86},
and poor correlation with the contrast between full noun phrases and
pronouns.
As noted in~\cite{passonneau&litman93},
the NP features largely reflect Passonneau's hypotheses
that adjacent utterances are more likely to contain expressions that
corefer, or that are inferentially linked, if they occur within the
same segment; and that a definite pronoun is more likely than a full
NP to refer to an entity that was mentioned in the current segment, if
not in the previous utterance.

\subsection{Evaluation}

The segmentation algorithms presented in the next two
sections were developed
by examining only a {\it training} set of narratives.  The algorithms
are then evaluated by examining their performance in predicting
segmentation on a separate {\it test} set.  We currently use 10
narratives for training and 5 narratives for testing.
(The remaining 5 narratives are reserved for future research.)
The 10 training
narratives
range in length from 51 to 162 phrases (Avg.=101.4), or from 38 to 121
clauses (Avg.=76.8).
The 5 test
narratives
range in length from 47 to 113 phrases (Avg.=87.4), or from
37 to 101 clauses  (Avg.=69.0).
The ratios of test to training data measured in
narratives, prosodic phrases and clauses, respectively, are 50.0\%, 43.1\% and
44.9\%.
For the machine learning algorithm we also estimate
performance using {\it
cross-validation}~\cite{KulikowskiBook90}, as detailed in
Section~\ref{mlearn}.

To quantify algorithm performance, we use the information
retrieval metrics shown in Fig.~\ref{ir}.
\begin{figure}[b]
{\scriptsize
\begin{center}
\begin{tabular}{| c | c  c | } \hline \hline
\multicolumn{1}{| c }{} & \multicolumn{2}{| c |}{Subjects}  \\\cline{2-3}
\multicolumn{1}{| c }{Algorithm} &  \multicolumn{1}{| c }{Boundary} &
\multicolumn{1}{ c|}{Non-Boundary} \\\hline
Boundary 	&\multicolumn{1}{c|}{a}
                &\multicolumn{1}{c|}{b}       \\\cline{2-3}
Non-Boundary    &\multicolumn{1}{c|}{c}
                &\multicolumn{1}{c|}{d}       \\\hline
\end{tabular}
\begin{tabbing}
aaaaaaaaaaaaaaaaaaaaaaaaa \= aaaaaaaaaaaaaaaaaaaaaaaaa \kill
\begin{minipage}{3in}{
{\small
\begin{description}
\item[Recall] = $\frac{a}{(a+c)}$
\item[Precision] = $\frac{a}{(a+b)}$
\end{description}}}
\end{minipage}
\>
\begin{minipage}{3in}{
{\small
\begin{description}
\item[Fallout] = $\frac{b}{(b+d)}$
\item[Error] = $\frac{(b+c)}{(a+b+c+d)}$
\end{description}}}
\end{minipage}
\end{tabbing}
\caption{\label{ir} Information retrieval metrics.}
\end{center}
}
\end{figure}
Recall is the ratio of correctly hypothesized boundaries to target
boundaries.
Precision is the ratio of hypothesized
boundaries that are correct to the total hypothesized boundaries.
(Cf. Fig.~\ref{ir} for fallout and error.)
Ideal behavior would be to identify all and only the target
boundaries: the values for
b and c in Fig.~\ref{ir} would thus both equal 0, representing no errors.
The ideal values for recall, precision, fallout, and error are 1, 1, 0, and 0,
while the worst values are 0, 0, 1, and 1.
To get an intuitive summary of overall performance,
we also sum the deviation of the observed value from the ideal value for each
metric:
(1-recall) $+$ (1-precision) $+$ fallout $+$ error.
The summed deviation for perfect performance is thus 0.

Finally, to interpret our quantitative results, we use the
performance of our human subjects as a target goal for the performance
of our algorithms~\cite{GCY}. Table~\ref{human} shows the average
human performance for both the training and test sets of narratives.
{\scriptsize
\begin{table}[t]
\begin{center}
\begin{tabular}{| l | r | r | r | r | r |}\hline
\multicolumn{1}{| c }{ } & \multicolumn{1}{| c }{Recall} &
\multicolumn{1}{| c }{Prec} & \multicolumn{1}{| c }{Fall} &
\multicolumn{1}{| c |}{Error} & \multicolumn{1}{| c |} {SumDev}\\\hline
Training Set       &.63   &.72   &.06   &.12   &.83	\\\hline
Test Set	       &.64   &.68   &.07   &.11   &.86		\\\hline
\end{tabular}
\end{center}
\caption{\label{human} Average human performance.}
\end{table}
}
Note that human performance is basically the same for both sets
of narratives. However, two factors prevent this performance from being closer
to ideal (e.g., recall and precision of 1).  The first is the wide
variation in the number of boundaries that subjects used, as discussed
above. The second is the inherently fuzzy nature of boundary location.
We discuss this second issue at length in~\cite{passonneau&litman95},
and present relaxed IR metrics that penalize near misses less heavily
in~\cite{Litman-Passonneau95a}.

\section{Hand Tuning}
\label{handTun}

To improve performance,
we analyzed the two types of IR errors
made by the original NP
algorithm~\cite{passonneau&litman93} on the training data.
Type ``b'' errors (cf. Fig.~\ref{ir}),
mis-classification of non-boundaries, were reduced by
changing the coding features pertaining to clauses and NPs.  Most ``b''
errors correlated with two conditions used in the NP algorithm,
identification of clauses and of inferential links.  The revision led to fewer
clauses (more assignments of `NA' for the three NP features) and more
inference relations.
One example of a change to clause coding is that formulaic
utterances having the structure of clauses, but which function like
interjections, are no longer recognized as independent clauses.
These include the phrases {\it let's see, let me see, I
don't know, you know} when they occur with no verb phrase argument.
Other changes pertained to sentence fragments, unexpected clausal
arguments, and embedded speech.

Three types of inference relations linking successive clauses
(C$_{i-1}$, C$_{i}$) were added (originally there were 5
types~\cite{passonneau94cod}).
Now, a pronoun (e.g., {\it it, that, this}) in C$_{i}$ referring to an
action, event or fact inferrable from C$_{i-1}$ links the two clauses.
So does an implicit argument, as in Fig.~\ref{ex1}, where the missing
argument of {\it notice} is inferred to be the event of the pears
falling.
The third case is where an NP in C$_{i}$ is described as
part of an event that results directly from an event mentioned in
C$_{i-1}$.

\begin{figure}[b]
{\scriptsize
\begin{tabular}{r r l}
\multicolumn{1}{c}{Cl.} &
\multicolumn{1}{c}{Phr.} & \\
6 & 3.01  & [1.1 [.7] A-nd] he's not really.. doesn't seem \\
  &       & \hspace{.18in}  to be paying all that much attention \\
7  &       & \hspace{.18in}[.55? because [.45]] you
   know {\it the pears fall}$_{i}$, \\
8 & 3.02  & and.. he doesn't really notice (\O$_{i}$), \\
\end{tabular}
\caption{\label{ex1} Inferential link due to implicit argument.}}
\end{figure}

``C'' type errors (cf. Fig.~\ref{ir}), mis-classification of boundaries,
often occurred where prosodic and cue features conflicted with NP
features.  The original NP algorithm assigned boundaries wherever the
three values `-coref', `-infer', `-global.pro' (defined in
section~\ref{method}) co-occurred, represented as the first conditional
statement of Fig.~\ref{np-alg}.  Experiments led to the hypothesis
that the most improvement came by assigning a boundary if the {\it
cue-prosody} feature had the value `complex', even if the algorithm
would not otherwise assign a boundary, as shown in Fig.~\ref{np-alg}.

{\scriptsize
\begin{figure}
\begin{tabbing}
ss\=ss\=ss\=ss\=ss\=ss\=ss\=ss\kill
{\bf if} (coref $=$ -coref {\bf and} infer $=$ -infer {\bf and}
    global.pro $=$ -global.pro) \\
\> \>{\bf then} {\it boundary} \\
\>{\bf elseif} cue-prosody $=$ complex {\bf then} {\it boundary} \\
{\bf else} {\it non-boundary}
\end{tabbing}
\caption{\label{np-alg} Condition 2 algorithm.}
\end{figure}}

{\scriptsize
\begin{table}[b]
\begin{center}
\begin{tabular}{| l | r | r | r | r | r |}\hline
\multicolumn{1}{| c }{Average} &
\multicolumn{1}{| c }{Recall} & \multicolumn{1}{| c }{Prec} &
\multicolumn{1}{| c }{Fall} & \multicolumn{1}{| c |}{Error} &
\multicolumn{1}{| c |}{SumDev}\\\hline
Condition 1  & .42 & .40 & .14 & .22 & 1.54 \\
Std. Dev.    & .17 & .12 & .06 & .07 & .34 \\ \hline
Condition 2  & .58 & .62 & .08 & .14 & 1.02 \\
Std. Dev.    & .14 & .10 & .04 & .05 & .18 \\  \hline
\end{tabular}
\caption{\label{tuned-np} Performance on training set.}
\end{center}
\begin{center}
\begin{tabular}{| l | r | r | r | r | r |}\hline
\multicolumn{1}{|c}{Average}
& \multicolumn{1}{| c }{Recall} & \multicolumn{1}{| c }{Prec} &
\multicolumn{1}{| c }{Fall} & \multicolumn{1}{| c |}{Error} &
\multicolumn{1}{| c |}{SumDev}\\\hline
Condition 1   & .44  & .29 & .16 & .21 & 1.64 \\
Std. Dev.     & .18  & .17 & .07 & .05 & .32 \\\hline
Condition 2  & .50 & .44 & .11 & .17 & 1.34  \\
Std. Dev.    & .21 & .06 & .03 & .04 &  .29 \\\hline
\end{tabular}
\caption{\label{test-set} Performance on test set.}
\end{center}
\end{table}}


We refer to the original NP algorithm applied to the initial coding as
Condition 1, and the tuned algorithm applied to the enriched coding
as Condition 2.  Table~\ref{tuned-np} presents the average IR scores
across the narratives in the {\it training} set for both conditions.
Reduction of ``b'' type errors
raises precision, and lowers fallout and error rate.  Reduction of ``c''
type errors raises recall, and lowers fallout and error rate.  All scores
improve in Condition 2, with precision and fallout
showing the greatest relative improvement.
The major difference from human performance is relatively poorer precision.

The standard
deviations in Table~\ref{tuned-np}
are often close to 1/4 or 1/3 of the reported averages.
This indicates a large amount
of variability in the data, reflecting wide differences across
narratives (speakers) in the training set
with respect to the distinctions recognized by the
algorithm.  Although the high standard
deviations show that the tuned algorithm is not well fitted to each
narrative, it is likely that it is overspecialized to the training sample
in the sense that test narratives are likely to exhibit further
variation.

Table~\ref{test-set} shows the results of the hand tuned algorithm on
the 5 randomly selected test narratives on both Conditions 1 and 2.
Condition 1 results, the untuned algorithm with the initial feature set,
are very similar to the training set except for worse precision.
Thus, despite the high standard deviations, 10 narratives seems to have
been a sufficient sample size for evaluating the initial NP algorithm.
Condition 2 results are better than condition 1 in Table~\ref{test-set},
and condition 1 in Table~\ref{tuned-np}.  This is strong
evidence that the tuned algorithm is a better predictor of segment
boundaries than the original NP algorithm.  Nevertheless, the test
results of condition 2 are much worse than the corresponding training results,
particularly for precision (.44 versus .62). This confirms that the
tuned algorithm is over calibrated to the training set.

\section{Machine Learning}
\label{mlearn}

We use the machine learning program C4.5~\cite{Quinlan93} to
automatically develop segmentation algorithms
from our corpus of coded
narratives, where each potential boundary site has been classified and
represented as a set of linguistic features.  The
first input to C4.5 specifies the names of the classes to be learned
({\it boundary} and {\it non-boundary}), and the names and potential
values of a fixed set of coding features (Fig.~\ref{features}).  The
second input is the training data, i.e., a set of examples for which
the class and feature values (as in Fig.~\ref{coding}) are
specified.  Our training set of 10 narratives provides 1004 examples
of potential boundary sites. The output of C4.5 is a classification
algorithm expressed as a decision tree, which predicts the class of a
potential boundary given its set of feature values.

Because machine learning makes it convenient to
induce decision trees under a wide variety of conditions, we
have performed numerous experiments, varying
the number of features used to code the training data, the definitions used for
classifying a potential boundary site as {\it boundary} or
{\it non-boundary}\footnote{\cite{Litman-Passonneau95a} varies
the number of subjects used to
determine boundaries.}
and the options available for running the C4.5 program.
Fig.~\ref{mlAlg} shows one of the highest-performing learned decision
trees from
our experiments.
\begin{figure}[t]
{\scriptsize
\begin{center}
\begin{tabbing}
ss\=ss\=ss\=ss\=ss\=ss\=ss\=ss\kill
{\bf if} before $=$ -sentence.final.contour {\bf then} {\it non-boundary} \\
{\bf elseif} before $=$ +sentence.final.contour {\bf then} \\
\>{\bf if}   coref $=$ NA {\bf then} {\it non-boundary} \\
\>{\bf elseif}   coref $=$ +coref {\bf then} \\
\>   \>{\bf if}   after $=$ +sentence.final.contour {\bf then}\\
\>   \>   \>{\bf if}   duration $\leq$ 1.3  {\bf then} {\it non-boundary} \\
\>   \>   \>{\bf elseif}   duration $>$ 1.3  {\bf then} {\it boundary} \\
\>   \>{\bf elseif}   after $=$ -sentence.final.contour {\bf then}\\
\>   \>   \>{\bf if}   word$_{1}$ $\in$
\{also,basically,because,finally,first,like,\\
\>  \> \> \>meanwhile,no,oh,okay,only,see,so,well,where,NA\}  \\
\>\>\>\>{\bf then} {\it non-boundary} \\
\>   \>   \>{\bf elseif}  word$_{1}$ $\in$ \{anyway,but,now,or,then\} {\bf
then} {\it boundary} \\
\>   \>   \>{\bf elseif} word$_{1}$ $=$ and {\bf then}\\
\>   \>   \>   \>{\bf if}   duration $\leq$ 0.6  {\bf then} {\it non-boundary}
\\
\>   \>   \>   \>{\bf elseif}   duration $>$ 0.6  {\bf then} {\it boundary} \\
\>{\bf elseif}   coref $=$ -coref {\bf then}\\
\>   \>{\bf if}   infer $=$ +infer {\bf then} {\it non-boundary} \\
\>   \>{\bf elseif}   infer $=$ NA {\bf then} {\it boundary} \\
\>   \>{\bf elseif}   infer $=$ -infer {\bf then}\\
\>   \>   \>{\bf if}   after $=$ -sentence.final.contour {\bf then} {\it
boundary} \\
\>   \>   \>{\bf elseif}   after $=$ +sentence.final.contour {\bf then}\\
\>   \>   \>   \>{\bf if}   cue$_{1}$ $=$ true {\bf then}\\
\>   \>   \>   \>   \>{\bf if}   global.pro $=$ NA {\bf then} {\it boundary} \\
\>   \>   \>   \>   \>{\bf elseif}   global.pro $=$ -global.pro {\bf then} {\it
boundary} \\
\>   \>   \>   \>   \>{\bf elseif}   global.pro $=$ +global.pro {\bf then}\\
\>   \>   \>   \>   \>   \>{\bf if}   duration $\leq$ 0.65  {\bf then} {\it
non-boundary} \\
\>   \>   \>   \>   \>   \>{\bf elseif}   duration $>$ 0.65  {\bf then} {\it
boundary} \\
\>   \>   \>   \>{\bf elseif}   cue$_{1}$ $=$ false {\bf then}\\
\>   \>   \>   \>   \>{\bf if}   duration $>$ 0.5  {\bf then} {\it
non-boundary} \\
\>   \>   \>   \>   \>{\bf elseif}   duration $\leq$ 0.5  {\bf then}\\
\>   \>   \>   \>   \>   \>{\bf if}   duration $\leq$ 0.35  {\bf then} {\it
non-boundary} \\
\>   \>   \>   \>   \>   \>{\bf elseif}   duration $>$ 0.35  {\bf then} {\it
boundary} \\
\end{tabbing}
\caption{\label{mlAlg} Learned decision tree for segmentation.}
\end{center}}
\end{figure}
This decision tree was learned under the following
conditions: all of the features shown in Fig.~\ref{features} were used
to code the training data, boundaries were classified as discussed in
section~\ref{method}, and C4.5 was run using only the default options.
The decision tree predicts the class of a potential boundary site
based on the features {\it before}, {\it after}, {\it duration}, {\it
cue$_{1}$}, {\it word$_{1}$}, {\it coref}, {\it infer}, and {\it
global.pro}.  Note that although not all available features are used
in the tree, the included features represent 3 of the 4 general types
of knowledge (prosody, cue phrases and noun phrases).  Each level of
the tree specifies a test on a single feature, with a branch for every
possible outcome of the test.\footnote{The actual tree branches on every
value of {\it word$_{1}$}; the figure merges these branches
for clarity.}   A branch can either lead to the
assignment of a class, or to another test.
For example, the tree
initially branches based on the value of the feature {\it before}.  If
the value is `-sentence.final.contour' then the first branch is taken
and the potential boundary site is assigned the class {\it
non-boundary}.  If the value of {\it before} is
`+sentence.final.contour' then the second branch is taken and the
feature {\it coref} is tested.

The performance of this learned decision tree averaged over the 10
training narratives is shown in Table~\ref{mlTrain}, on the line
labeled ``Learning 1''.  The line labeled ``Learning 2'' shows the results
from
another machine learning experiment, in which one of the default
C4.5 options used in ``Learning 1'' is overridden.
The ``Learning 2'' tree (not shown due to space restrictions) is more
complex than the tree of Fig.~\ref{mlAlg}, but has slightly better
performance.  Note that ``Learning 1'' performance
is comparable to human performance (Table~\ref{human}), while ``Learning 2''
is slightly better than humans.
The results obtained via
machine learning are also somewhat better than the results obtained
using hand tuning---particularly with respect to precision (``Condition 2''
in Table~\ref{tuned-np}), and are a great improvement over the original
NP results (``Condition 1'' in Table~\ref{tuned-np}).

The performance of the learned decision trees averaged over the 5 test
narratives is shown in Table~\ref{mlTest}.  Comparison of
Tables~\ref{mlTrain} and~\ref{mlTest} shows that, as with the hand
tuning results (and as expected), average performance is worse when
applied to the testing rather than the training data particularly with
respect to precision.  However, performance is an improvement over our
previous best
results (``Condition 1'' in
Table~\ref{test-set}),
and is comparable to (``Learning 1'') or very slightly
better than (``Learning 2'') the hand tuning results (``Condition 2'' in
Table~\ref{test-set}).

We also use the resampling method of {\it
cross-validation}~\cite{KulikowskiBook90} to estimate performance, which
averages results over multiple
partitions of a sample into test versus training data.
We performed 10 runs of the learning program, each using 9 of the 10
training narratives for that run's training set (for learning the
tree) and the remaining narrative for testing.  Note that for {\it
each} iteration of the cross-validation, the learning process begins
from scratch and thus each training and testing set are still
disjoint.  While this method does not make sense for humans, computers
can truly ignore previous iterations.
For sample sizes in the hundreds (our
10 narratives provide 1004 examples) 10-fold cross-validation often
provides a better performance estimate than the hold-out
method~\cite{KulikowskiBook90}.
Results using cross-validation are shown in Table~\ref{mlXval}, and are
better than the estimates obtained using the hold-out
method (Table~\ref{mlTest}), with the major improvement coming
from precision.  Because a different tree is learned on
each iteration, the cross-validation evaluates the learning method,
not a particular decision tree.

{\scriptsize
\begin{table}
\begin{center}
\begin{tabular}{| l | r | r | r | r | r |}\hline
\multicolumn{1}{| c }{Average} & \multicolumn{1}{| c }{Recall} &
\multicolumn{1}{| c }{Prec} & \multicolumn{1}{| c }{Fall} &
\multicolumn{1}{| c |}{Error} & \multicolumn{1}{| c |} {SumDev}\\\hline
Learning 1    &.54   &.76   &.04   &.11   &.85        \\
Std. Dev. & .18 & .12 & .02 & .04 & .28 \\\hline
Learning 2    &.59   &.78   &.03   &.10   &.76	\\
Std. Dev. & .22 & .12 & .02 & .04 & .29\\\hline
\end{tabular}
\caption{\label{mlTrain} Performance on training set.}
\end{center}
\begin{center}
\begin{tabular}{| l | r | r | r | r | r |}\hline
\multicolumn{1}{| c }{Average } & \multicolumn{1}{| c }{Recall} &
\multicolumn{1}{| c }{Prec} & \multicolumn{1}{| c }{Fall} &
\multicolumn{1}{| c |}{Error} & \multicolumn{1}{| c |} {SumDev}\\\hline
Learning 1    &.43   &.48   &.08   &.16   &1.34       \\
Std. Dev.& .21 & .13 & .03 & .05 & .36\\\hline
Learning 2	&.47   &.50   &.09   &.16   &1.27	\\
Std. Dev.& .18  & .16 & .04 & .07 & .42\\\hline
\end{tabular}
\caption{\label{mlTest} Performance on test set.}
\end{center}
\begin{center}
\begin{tabular}{| l | r | r | r | r | r |}\hline
\multicolumn{1}{| c }{Average } & \multicolumn{1}{| c }{Recall} &
\multicolumn{1}{| c }{Prec} & \multicolumn{1}{| c }{Fall} &
\multicolumn{1}{| c |}{Error} & \multicolumn{1}{| c |} {SumDev}\\\hline
Learning 1    &.43  &.63  &.05  &.15  &1.14 \\
Std. Dev.& .19 & .16 & .03 & .03 & .24 \\\hline
Learning 2	&.46  &.61  &.07  &.15  &1.15 \\
Std. Dev.& .20 & .14 & .04 & .03 & .21 \\\hline
\end{tabular}
\caption{\label{mlXval} Using 10-fold cross-validation.}
\end{center}
\end{table}
}

\section{Conclusion}
\label{conclusion}

We have presented two methods for developing segmentation
hypotheses using multiple linguistic features.  The first method
hand tunes features and algorithms
based on analysis of training errors.
The second method,
machine learning, {\it automatically} induces decision trees from
coded corpora.
Both methods rely on an
enriched set of input features compared to our previous work.  With
each method, we have achieved marked improvements in performance
compared to our previous work and are approaching human performance.
Note that quantitatively, the machine learning results are slightly
better than the hand tuning results.  The main difference on average
performance is the higher precision of the automated algorithm.
Furthermore, note that the machine learning algorithm used the changes
to the coding features that resulted from the tuning methods.  This
suggests that hand tuning is a useful method for understanding how to
best code the data, while machine learning provides an effective (and
automatic) way to produce an algorithm given a good feature
representation.

Our results lend further support to the hypothesis that linguistic devices
correlate with discourse structure (cf.  section~\ref{relWk}), which
itself has practical import.  Understanding systems could
infer segments as a step towards producing summaries,
while generation systems could signal segments to increase
comprehensibility.\footnote{Cf.~\cite{hirschberg&pierrehumbert86} who
argue that
comprehensibility improves if units are prosodically signaled.}
Our results also suggest that to best identify or convey segment
boundaries, systems will need to exploit {\it multiple} signals
simultaneously.

We plan to continue our experiments by further merging the automated
and analytic techniques, and evaluating new algorithms on our final test
corpus.  Because we have already used cross-validation, we do not
anticipate significant degradation on new test narratives.  An important
area for future research is to develop principled methods
for identifying distinct speaker strategies pertaining to
how they signal segments.  Performance of
individual speakers varies widely as shown by the high standard
deviations in our tables.  The original NP,
hand tuned, and machine learning algorithms all do relatively
poorly on narrative 16 and relatively well on 11 (both in the test set)
under all conditions.
This lends support to the hypothesis that there may be consistent
differences among speakers regarding strategies for
signaling shifts in global discourse structure.

\bibliographystyle{acl}

\begin{thebibliography}{}

\bibitem[\protect\citename{Breiman \bgroup et al.\egroup }1984]{brieman84}
Leo Breiman, Jerome Friedman, Richard Olshen, and C.~Stone.
\newblock 1984.
\newblock {\em Classification and Regression Trees}.
\newblock Wadsworth and Brooks, Monterey, CA.

\bibitem[\protect\citename{Chafe}1980]{chafe80}
Wallace~L. Chafe.
\newblock 1980.
\newblock {\em The Pear Stories}.
\newblock Ablex Publishing Corporation, Norwood, NJ.

\bibitem[\protect\citename{Gale \bgroup et al.\egroup }1992]{GCY}
William Gale, Ken~W. Church, and David Yarowsky.
\newblock 1992.
\newblock Estimating upper and lower bounds on the performance of word-sense
  disambiguation programs.
\newblock In {\em Proc. of the 30th ACL}, pages 249--256.

\bibitem[\protect\citename{Grosz and Hirschberg}1992]{grosz&hirschberg92}
Barbara Grosz and Julia Hirschberg.
\newblock 1992.
\newblock Some intonational characteristics of discourse structure.
\newblock In {\em Proc. of the International Conference on Spoken Language
  Processing}.

\bibitem[\protect\citename{Grosz and Sidner}1986]{gs86}
Barbara Grosz and Candace Sidner.
\newblock 1986.
\newblock Attention, intentions and the structure of discourse.
\newblock {\em Computational Linguistics}, 12:175--204.

\bibitem[\protect\citename{Grosz \bgroup et al.\egroup }1983]{gjw83}
Barbara~J. Grosz, Aaravind~K. Joshi, and Scott Weinstein.
\newblock 1983.
\newblock Providing a unified account of definite noun phrases in discourse.
\newblock In {\em Proc. of the 21st ACL}, pages 44--50.

\bibitem[\protect\citename{Hearst}1994]{hearst94}
Marti~A. Hearst.
\newblock 1994.
\newblock Multi-paragraph segmentation of expository text.
\newblock In {\em Proc. of the 32nd ACL}.

\bibitem[\protect\citename{Hirschberg and Grosz}1992]{hirschberg&grosz92}
Julia Hirschberg and Barbara Grosz.
\newblock 1992.
\newblock Intonational features of local and global discourse structure.
\newblock In {\em Proc. of the Darpa Workshop on Spoken Language}.

\bibitem[\protect\citename{Hirschberg and Litman}1993]{hl93}
Julia Hirschberg and Diane Litman.
\newblock 1993.
\newblock Empirical studies on the disambiguation of cue phrases.
\newblock {\em Computational Linguistics}, 19(3):501--530.

\bibitem[\protect\citename{Hirschberg and
  Pierrehumbert}1986]{hirschberg&pierrehumbert86}
Julia Hirschberg and Janet Pierrehumbert.
\newblock 1986.
\newblock The intonational structuring of discourse.
\newblock In {\em Proc. of the 24th ACL}.

\bibitem[\protect\citename{Hobbs}1979]{hobbs79}
Jerry~R. Hobbs.
\newblock 1979.
\newblock Coherence and coreference.
\newblock {\em Cognitive Science}, 3(1):67--90.

\bibitem[\protect\citename{Isard and Carletta}1995]{Isard&Carletta95}
Amy Isard and Jean Carletta.
\newblock 1995.
\newblock Replicability of transaction and action coding in the map task
  corpus.
\newblock In {\em AAAI 1995 Spring Symposium Series: Empirical Methods in
  Discourse Interpretation and Generation}, pages 60--66.

\bibitem[\protect\citename{Kameyama}1986]{kameyama86}
Megumi Kameyama.
\newblock 1986.
\newblock A property-sharing constraint in centering.
\newblock In {\em Proc. of the 24th ACL}, pages 200--206.

\bibitem[\protect\citename{Kozima}1993]{kozima93}
H.~Kozima.
\newblock 1993.
\newblock Text segmentation based on similarity between words.
\newblock In {\em Proc. of the 31st ACL (Student Session)}, pages 286--288.

\bibitem[\protect\citename{Lascarides and
  Oberlander}1992]{lascarides&oberlander92}
Alex Lascarides and Jon Oberlander.
\newblock 1992.
\newblock Temporal coherence and defeasible knowledge.
\newblock {\em Theoretical Linguistics}.

\bibitem[\protect\citename{Linde}1979]{linde79}
Charlotte Linde.
\newblock 1979.
\newblock Focus of attention and the choice of pronouns in discourse.
\newblock In Talmy Givon, editor, {\em Syntax and Semantics: Discourse and
  Syntax}, pages 337--354. Academic Press, New York.

\bibitem[\protect\citename{Litman and Passonneau}1995]{Litman-Passonneau95a}
Diane~J. Litman and Rebecca~J. Passonneau.
\newblock 1995.
\newblock Developing algorithms for discourse segmentation.
\newblock In {\em AAAI 1995 Spring Symposium Series: Empirical Methods in
  Discourse Interpretation and Generation}, pages 85--91.

\bibitem[\protect\citename{Litman}1994]{litman94}
Diane~J. Litman.
\newblock 1994.
\newblock Classifying cue phrases in text and speech using machine learning.
\newblock In {\em Proc. of the 12th AAAI}, pages 806--813.

\bibitem[\protect\citename{Mann and Thompson}1988]{mann&thompson88}
William~C. Mann and Sandra Thompson.
\newblock 1988.
\newblock Rhetorical structure theory.
\newblock {\em TEXT}, pages 243--281.

\bibitem[\protect\citename{Moore and Paris}1993]{moore&paris93}
Johanna~D. Moore and Cecile Paris.
\newblock 1993.
\newblock Planning text for advisory dialogues: Capturing intentional and
  rhetorical information.
\newblock {\em Computational Linguistics}, 19:652--694.

\bibitem[\protect\citename{Moore and Pollack}1992]{moore&pollack92}
Johanna~D. Moore and Martha~E. Pollack.
\newblock 1992.
\newblock A problem for {RST}: The need for multi-level discourse analysis.
\newblock {\em Computational Linguistics}, 18:537--544.

\bibitem[\protect\citename{Morris and Hirst}1991]{morris&hirst91}
Jane Morris and Graeme Hirst.
\newblock 1991.
\newblock Lexical cohesion computed by thesaural relations as an indicator of
  the structure of text.
\newblock {\em Computational Linguistics}, 17:21--48.

\bibitem[\protect\citename{Moser and Moore}1995]{Moser&Moore95}
Megan Moser and Julia~D. Moore.
\newblock 1995.
\newblock Using discourse analysis and automatic text generation to study
  discourse cue usage.
\newblock In {\em AAAI 1995 Spring Symposium Series: Empirical Methods in
  Discourse Interpretation and Generation}, pages 92--98.

\bibitem[\protect\citename{Nakatani \bgroup et al.\egroup }1995]{Nakatani95}
Christine~H. Nakatani, Julia Hirschberg, and Barbara~J. Grosz.
\newblock 1995.
\newblock Discourse structure in spoken language: Studies on speech corpora.
\newblock In {\em AAAI 1995 Spring Symposium Series: Empirical Methods in
  Discourse Interpretation and Generation}, pages 106--112.

\bibitem[\protect\citename{Passonneau and Litman}1993]{passonneau&litman93}
Rebecca~J. Passonneau and Diane~J. Litman.
\newblock 1993.
\newblock Intention-based segmentation: Human reliability and correlation with
  linguistic cues.
\newblock In {\em Proc. of the 31st ACL}, pages 148--155.

\bibitem[\protect\citename{Passonneau and Litman}to
  appear]{passonneau&litman95}
Rebecca~J. Passonneau and D.~Litman.
\newblock to appear.
\newblock Empirical analysis of three dimensions of spoken discourse.
\newblock In E.~Hovy and D.~Scott, editors, {\em Interdisciplinary Perspectives
  on Discourse}. Springer Verlag, Berlin.

\bibitem[\protect\citename{Passonneau}1994]{passonneau94cod}
Rebecca~J. Passonneau.
\newblock 1994.
\newblock Protocol for coding discourse referential noun phrases and their
  antecedents.
\newblock Technical report, Columbia University.

\bibitem[\protect\citename{Passonneau}to appear]{passonneau95}
Rebecca~J. Passonneau.
\newblock to appear.
\newblock Interaction of the segmental structure of discourse with explicitness
  of discourse anaphora.
\newblock In E.~Prince, A.~Joshi, and M.~Walker, editors, {\em Proc. of the
  Workshop on Centering Theory in Naturally Occurring Discourse}. Oxford
  University Press.

\bibitem[\protect\citename{Polanyi}1988]{polanyi88}
Livya Polanyi.
\newblock 1988.
\newblock A formal model of discourse structure.
\newblock {\em Journal of Pragmatics}, pages 601--638.

\bibitem[\protect\citename{Quinlan}1993]{Quinlan93}
John~R. Quinlan.
\newblock 1993.
\newblock {\em C4.5 : Programs for Machine Learning}.
\newblock Morgan Kaufmann Publishers, San Mateo, Calif.

\bibitem[\protect\citename{Reichman}1985]{Reichman85}
Rachel Reichman.
\newblock 1985.
\newblock {\em Getting Computers to Talk Like You and Me: Discourse Context,
  Focus, and Semantics}.
\newblock Bradford. MIT, Cambridge.

\bibitem[\protect\citename{Reynar}1994]{Reynar94}
J.~C. Reynar.
\newblock 1994.
\newblock An automatic method of finding topic boundaries.
\newblock In {\em Proc. of the 32nd ACL (Student Session)}, pages 331--333.

\bibitem[\protect\citename{Stifleman}1995]{stifleman95}
Lisa~J. Stifleman.
\newblock 1995.
\newblock A discourse analysis approach to structured speech.
\newblock In {\em AAAI 1995 Spring Symposium Series: Empirical Methods in
  Discourse Interpretation and Generation}, pages 162--167.

\bibitem[\protect\citename{Webber}1991]{webber91}
Bonnie~L. Webber.
\newblock 1991.
\newblock Structure and ostension in the interpretation of discourse deixis.
\newblock {\em Language and Cognitive Processes}, pages 107--135.

\bibitem[\protect\citename{Weiss and Kulikowski}1991]{KulikowskiBook90}
Sholom~M. Weiss and Casimir Kulikowski.
\newblock 1991.
\newblock {\em Computer systems that learn: classification and prediction
  methods from statistics, neural nets, machine learning, and expert systems}.
\newblock Morgan Kaufmann.

\end{thebibliography}

\end{document}